%% file: MAIN.tex
\pdfoutput=1
\documentclass[11pt]{article}

\usepackage[final]{acl}

\usepackage{times}
\usepackage{latexsym}
\usepackage{listings}
\usepackage{textcomp}
\usepackage{amssymb}
\usepackage{amsmath}
\usepackage{upquote}

\usepackage[table]{xcolor}

\usepackage[T1]{fontenc}

\usepackage[utf8]{inputenc}

\usepackage{microtype}

\usepackage{inconsolata}

\usepackage{graphicx}
\usepackage{stfloats}
\usepackage{tabularx,array,enumitem}
\newcolumntype{Y}{>{\raggedright\arraybackslash}X}

\input{COMMANDS}

\title{so much depends / upon / a whitespace:\newline Why Whitespace Matters for Poets and LLMs}

\newcommand{\aspace}{\hspace{1em}}
\newcommand{\uw}{$^\diamondsuit$}
\newcommand{\cu}{$^\spadesuit$}

\author{Sriharsh Bhyravajjula\uw \aspace Melanie Walsh\uw \aspace Anna Preus\uw \aspace Maria Antoniak\cu 
\vspace{.3em}\\
\small{\uw University of Washington \aspace \cu University of Colorado Boulder}}

\begin{document}

\maketitle

\begin{abstract}
Whitespace is a critical component of poetic form, reflecting both adherence to standardized forms and rebellion against those forms.
Each poem's whitespace distribution reflects the artistic choices of the poet and is an integral semantic and spatial feature of the poem.
Yet, despite the popularity of poetry as both a long-standing art form and as a generation task for large language models (LLMs), whitespace has not received sufficient attention from the NLP community.
Using a corpus of 19k English-language published poems from Poetry Foundation, we investigate how 4k poets have used whitespace in their works.  We release a subset of 2.8k public-domain poems with preserved formatting to facilitate further research in this area. We compare whitespace usage in the published poems to (1) 51k LLM-generated poems, and (2) 12k unpublished poems posted in an online community. We also explore whitespace usage across time periods, poetic forms, and data sources. Additionally, we find that different text processing methods can result in significantly different representations of whitespace in poetry data, motivating us to use these poems and whitespace patterns to discuss implications for the processing strategies used to assemble pretraining datasets for LLMs.
\end{abstract}

\section{Introduction}

For many text datasets, whitespace is treated as a minor concern, not critical to a text's meaning.
It is often standardized or stripped before further processing.
But in poetry, whitespace matters.
It is vitally important, perhaps even the most defining feature of the genre (at least on the page).
\citet{van_dijk_reading_2011} argues that ``there is only one characteristic which immediately distinguishes modern poetry from prose: the blank space surrounding the text.''
In poetry, whitespace---including line and stanza breaks, indentation, space between words, and more---is not merely stylistic flair but integral to structure, meaning, and the reading experience.

\begin{figure}[t]
    \includegraphics[width=7.5cm]{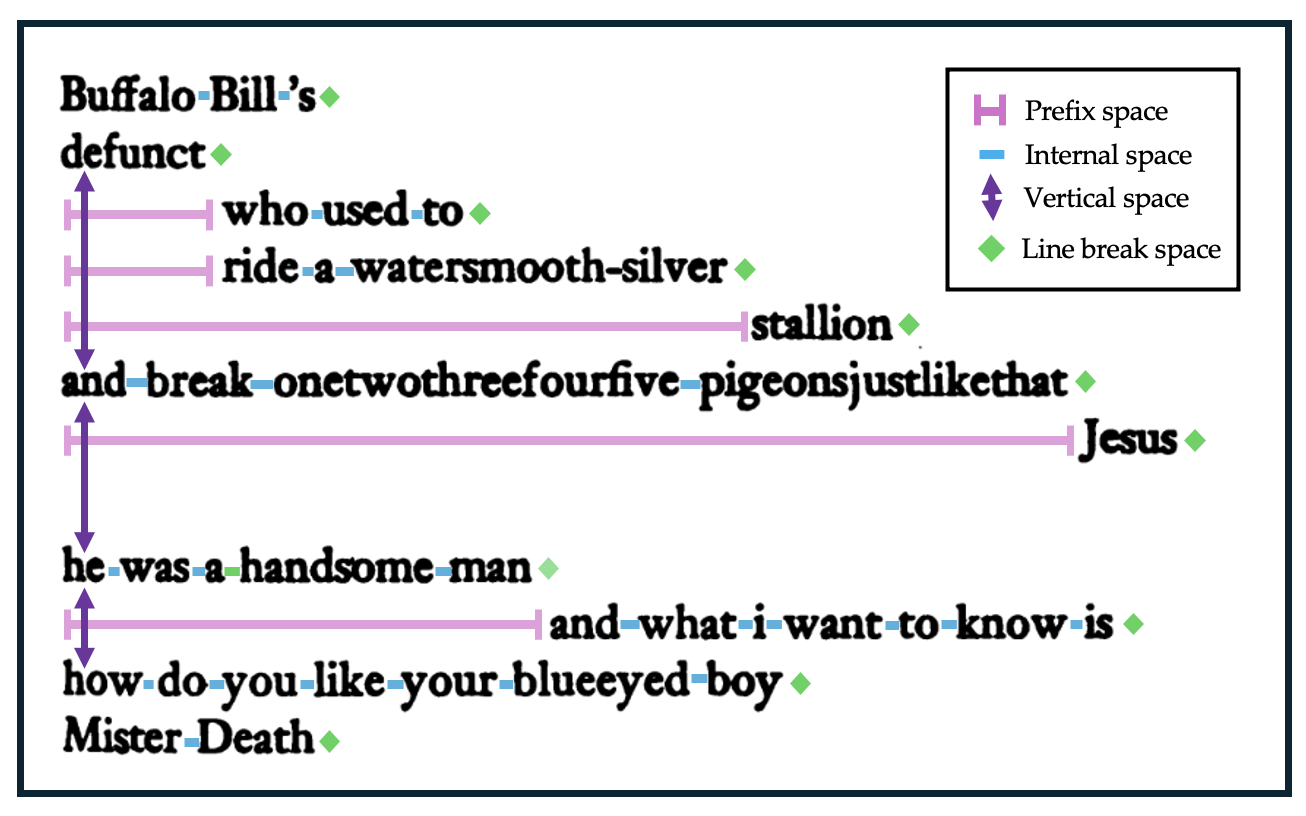}
    \caption{An excerpt from ``[Buffalo Bill 's]'' (1926) by E.E. Cummings, annotated using our whitespace typology, WISP\feather{} (Whitespace In Spatial Poetics). WISP distinguishes between five categories of whitespace usage: \colorbox{break-color}{\texttt{line breaks}}, \colorbox{prefix-color}{\texttt{prefix}} space, \colorbox{internal-color}{\texttt{internal}} space, \colorbox{vertical-color}{\texttt{vertical}} space, and \colorbox{length-color}{\texttt{line length}}.}
    \label{figure:fig1}
\end{figure}

Yet whitespace has received relatively little attention in NLP. 
One reason is that it is often deemed inconsequential. Another is that it is technically challenging to represent and preserve. 
There are over 25 different Unicode characters that encode whitespace of varying widths and functions. 
On the web, whitespace is often represented with HTML and CSS styling---a difficult task in its own right, and one that also poses problems for converted plain text formatting. 
What's more, with poetry, it’s not always possible to tell whether a line break or other whitespace reflects the author’s intent, the original typesetting, or an artifact of reprinting or digitization.
In the digital humanities (DH), scholars studying poetry often painstakingly encode layouts using the XML-based TEI (Text Encoding Initiative),\footnote{https://www.tei-c.org} which underscores how central---and labor-intensive---whitespace preservation can be.

Whitespace not only has consequences for poetry but for NLP more broadly.
For LLMs, it turns out, whitespace also matters.
Research has shown that some models fail to account for the visual dimensions of text, and that adversarial attacks can exploit LLM challenges with vertical and horizontal space \cite{liVulnerabilityLLMsVertically2025, caiEvadeChatGPTDetectors2023}. 
Recent efforts have begun to address these and other issues by trying to preserve structure and whitespace during pretraining data preparation, highlighting growing awareness that layout carries meaning, especially for programming and mathematical data \cite{paster2023openwebmathopendatasethighquality, overwijk2022clueweb2210billionweb}.

In this work, we conduct a large computational study of poetry focused on whitespace: both its poetic implications and its challenges for LLMs.

Our contributions include:
\vspace{-0.5em}
\begin{itemize}
    \itemsep0em
    \item a large-scale analysis of whitespace usage across 19k published, 12k unpublished, and 51k generated poems (all in English),
    \item a focused whitespace usage analysis of published poems across 500 years, 4k poets, and diverse poetic forms,
    \item a dataset of 2.8k public-domain poems with preserved formatting to facilitate further research in this area,\footnote{Our data, code, and an interactive dashboard are available at \href{https://github.com/darthbhyrava/wisp}{https://github.com/darthbhyrava/wisp}} 
    \item the Whitespace In Spatial Poetics (WISP\feather{}) typology, that unifies different poetic studies of whitespace,
    \item and evaluation of various pretraining linearization systems, including established methods as well as experimental systems using multimodal models, leading to reflections on whitespace handling for LLMs.
\end{itemize}

\section{Related Work}

\subsection{Humanities Studies of Whitespace}

Whitespace is an important textual element of poetry \cite{drucker_graphical_2006, brinkman_making_2009, van_dijk_reading_2011, johnston_theorizing_2010} and a key expressive tool for poets, especially modern and contemporary poets \cite{halter_poem_2015, peterson_readable_1995, rollison_poem_2003, drucker_visible_1994}. The term ``white space'' emerged in English print in the late 1800s, referring to ``the blank areas of a page or other piece of printed matter,'' which were often ``regarded collectively as an element of layout and design'' \cite{OED_2025}. The \textit{Oxford English Dictionary} notes the term's shift to a single word with the rise of computing, where it is used to indicate ``blank space in electronic text produced by one or more keyed characters, as spaces, tabs, line-breaks, etc''  \cite{OED_2025}. We use the one-word term to refer to typographic whitespace in both print and digital contexts. 

    Long before either version of the term was coined, whitespace performed important functions in written verse, visually signaling the division of poems into lines and line groups. The line is a fundamental concept in poetry \cite{Van_der_Zee_2011}.
    While the poetic line is not defined by its visual representation on the page, practically, lines are indicated through whitespace that surrounds them.
    
    Conceptions of the poetic line, poetic form, and the relationship between poems and their visual representation on the page are also historically and culturally specific \cite{Prins_2008, Martin_2012, Jackson_2023}. In early modern and 18th-century English poetry, the division of poetic texts into lines generally corresponded to repeated patterns of sound, specifically patterns of meter and rhyme \cite{Fussell_1965, Brogan_1981, Attridge_1982, Martin_2012}. 
    With the rise of free verse in the late 19th and early 20th centuries \cite{Hartman_1980, Finch_2000, Beyers_2001}, however, lines were no longer necessarily defined by such metrical patterns, and 
    many poets began influentially incorporating new and varied forms of lineation into their poetry \cite{Hollander_1975, Berry_1989, peterson_readable_1995, Gross_1996, Beyers_2001, johnston_theorizing_2010}.
    
    Within this context, whitespace became an increasingly important expressive tool for poets, who incorporated variant spacing within and between their poetic lines and experimented with boundary-pushing typography and layouts \cite{perloff_futurist_1986,mcgann_black_1993, brinkman_making_2009, cundy_marinetti_1981}. These expressive usages of whitespace are important interpretive aspects of poems. In digital and digitized texts, however, standard and expressive uses of whitespace in poetry may be encoded in a range of ways, and their particularities can get flattened through various technical processes.

\subsection{Computational Studies of Whitespace}

In the digital humanities, studies of whitepace in poetry have focused on line breaks and enjambment.
Most closely related to our work is a study of enjambment by \citet{ruiz-etal-2017-enjambment}, which considers different types of enjambment in a small dataset of 3.7K Spanish-language sonnets.
Using hand-crafted rules and constituency and dependency parses, they detected the presence and type of enjambment and provided a visualization of which line position are more likely to contain enjambments.
Similar work on very small datasets (e.g., $N=69$) has used audio files as well as syntactic analysis to study types of enjambment \citep{hussein2018automatic}.
\citet{monget2020computational} provides a useful overview of this prior work on computational analyses of enjambment.

Prior work in NLP has mostly treated all whitespace uniformly. 
The places where whitespace has been seriously considered have mostly been (1) language-specific tokenization \citep{wiechetek-etal-2019-seeing} and (2) correction of OCR errors \citep{soni-etal-2019-correcting,bast-etal-2023-fast}.
However, there has been recent attention to whitespace formatting in pretraining datasets dedicated to programming and mathematical datasets and tasks \citep{paster2023openwebmathopendatasethighquality}.

Standard processes of macrodata refinement include quality filtering, removal of ``junk'' text, and tokenization. 
A critical but sometimes overlooked step is \textit{linearization}, the process by which web scraped data is transformed from HTML to text ready for use in pretraining \citep{dolma}.
Commercial tools exist to support this process, but while some comparisons have been done \citep{li2025datacomplmsearchgenerationtraining}, overall the research community has focused on the (also important) effects of quality filters \citep{lucy-etal-2024-aboutme}, curation \citep{wettig2025organizewebconstructingdomains}, and tokenization \citep{ali-etal-2024-tokenizer,wang2025tokenizationmattersdegradinglarge,whittington2024tokenisationnpcomplete,singh2024tokenizationcountsimpacttokenization,zheng2025brokentokenslanguagemodel}, where whitespace is usually treated as a separator rather than a feature with its own importance.

\section{WISP\feather{}: A Whitespace Typology}

\input{tables/typology}

Poets use whitespace in a variety of ways, both as a standard feature of traditional poetic forms (via line or stanza breaks) and as a more idiosyncratic, expressive tool (e.g., through inline spacing, indentation, or irregular line lengths).
To formalize these usages, we develop a practical typology of whitespace usage categories: the Whitespace In Spatial Poetics (WISP\feather{}) typology.
These categories can be combined, repeated, interjected, and used for larger patterns to shape the visual structure of a poem.
An overview of the typology is in Table \ref{table:typology}.

\paragraph{\colorbox{break-color}{Line Breaks}}

Line breaks refer to spaces that mark the end of a line of text, affecting the length, position, metrical composition, and rhythmic qualities of poetic lines \cite{Beyers_2001, Rosko_Zee_2011, Hazelton_2014}. Line breaks correspondingly hold significant weight for poets \cite{Levertov_1979, Fagan_2011, halter_poem_2015}, often  marking the place for rhymes, and in many ways defining the relationship between line and syntax \cite{longenbach2008poeticline}. Line breaks may come at the end of  sentences or syntactic units or they may fragment these units, carrying words, phrases, clauses, or sentences across vertical space. 

\paragraph{\colorbox{prefix-color}{Line Prefix Space}}
Line prefix spaces refer to instances of leading whitespace before a line, which introduce indentation from the left-margin. Many usages of prefix spaces in printed poetry are fairly standardized. As \citet{Ruwet_2014} notes, ``The width of the left margin is generally uniform, though the beginnings of some lines may be indented, often at regular intervals.'' \citet{Jacobson_2008} and \citet{Pacheco_2006} explore conventions for poetry publication through early printers' manuals, discussing a number of different conventional uses of indentation, including at the beginning of stanzas and to align pairs of rhymed lines. Prefix spacing can also be used in more unconventional ways \cite{Matore_2024, drucker_graphical_2006}, however, moving beyond traditional indentation to break up the text more radically, as in the poem in Figure~\ref{figure:fig1}. 

\paragraph{\colorbox{internal-color}{Internal Space}}

Internal space refers to non-standard whitespace that occurs within lines, appearing  between or within words. With the shift toward more focus on the visual elements of poetry \cite{van_dijk_reading_2011}, use of internal space within lines became more important. In her work on letterpress printing, \citet{Drucker_1984} notes that ``Writing produces a visual image: the shapes, sizes and placement of letters on a page contribute to the message produced, creating statements which cannot always be rendered in spoken language.'' The use of internal spacing can create this kind of visual feature, and also has other potential effects, including indicating a pause, breaking up a semantic unit, or contributing to broader visual patterning.

\paragraph{\colorbox{vertical-color}{Vertical Space}}

Vertical space refers to blank spaces between lines of text, which in digital poems are created through at least two line breaks, which create one or more lines of whitespace between lines of text. In conventional poetry printing, these blank lines were generally used to separate stanzas and line groups \cite{Jacobson_2008}. However, they can also be used in more unconventional and expressive ways. Writing about modern poetic forms in his influential ``Lecture on Modern Poetry,'' \citet{Hulme_1955}, suggests that the ``new verse resembles sculpture rather than music,'' arguing that ``it appeals to the eye rather than to the ear.'' Vertical spacing is a key element of this kind of sculptural poetry as well as a standard way of dividing up groups of poetic lines.

\paragraph{\colorbox{length-color}{Line Lengths}}

Line length refers to the number of visible characters or words in a poetic line. The length of a poetic line may be defined by patterns of sound or by visual choices and in either case it holds poetic meaning \cite{rosko_broken_2011}. \citet{Hollander_1975} highlights changes in common lengths of poetic lines in American poetry with the rise of free verse, suggesting, there is ``a widespread, received, free-verse style marked by a narrow (25-30 em) format, strong use of line-ending as a syntactical marker, etc., which plays about the same role in the ascent to paradise as the received Longfellow style did a century ago.'' This favor for short, sculptural lines is often associated with 20th-century poets like William Carlos Williams, who \citet{Dolin_1993} argues, ``created a revolution in poetic form'' by emphasizing ``the visual properties of the line,'' especially via concision.

\section{Data}

We collect three sources of English-language poetry data for comparison: published poems featured on the Poetry Foundation's website, unpublished poems shared in an online community, and LLM-generated poems.
We provide an overview of these datasets and their sizes in Table~\ref{table:data}.

\input{tables/data}

\subsection{Unpublished Poems \reddit{} }
\label{subsection:data-reddit}

We gather 12k poems from \texttt{r/OCPoetry} using ConvoKit \cite{chang-etal-2020-convokit}.
Most of these poems are not tagged with their form by the poet, so we automatically tag each poem with a form using the prompting framework from \citet{walsh-etal-2024-sonnet}, which reported high precision and recall for free-verse using GPT-4. 
Using this method with GPT 4.1,\footnote{\texttt{gpt-4.1-2025-04-14}} we identify 7,862 free-verse poems, 2,234 quatrains, 1,237 couplets, 608 tercets, and a smaller number of other forms.
These form labels allow us to directly compare whitespace usage in free-verse poems across data sources.

\subsection{LLM-Generated Poems \sparkle{}}

We use GPT-4 (OpenAI) and Sonnet 3.7 (Anthropic) to generate new datasets of poems.\footnote{\texttt{gpt-4, claude-3-7-sonnet-20250219}}
To generate poems on diverse themes, we randomly sample one poem for each poet in our Poetry Foundation dataset, resulting in 4,330 poems that we use as seeds whose title and poet are inserted in the prompt, generating three new poems per seed poem.
We use two prompt variations: (1) a prompt that explicitly requests the model to use whitespace creatively and (2) a simplified prompt that does not mention whitespace (see Appendix \ref{appendix-poem-generation-prompt}). 
Manual examination of the generated poems and explanations reveals that they are nearly all free verse, and so we use these poems in comparison only to  free verse poems from Poetry Foundation and Reddit.

\subsection{Published Poems \book{}}
\label{subsection:data-published-poems}

We scrape 19k poems from the public website of the Poetry Foundation, a U.S.-based nonprofit organization that amplifies poetry for a global audience through grants, awards, fellowships, digital outreach, and publication of the \textit{Poetry} magazine.

All the poems we analyze are freely available online, but some of the poems are in-copyright.
To ensure responsible data sharing, we release only the subset of poems that are in the public domain. 
In the U.S., as of 2025, works published in or before 1929 have entered the public domain. 
We share poems that were published in or before 1929, poems whose authors died in or before 1929, and poems that are explicitly marked as in the public domain.

We follow the methods described in \citet{walsh-etal-2024-sonnet} to measure how many of the poems appear in the Dolma pretraining dataset and how may of the poems were likely seen by a large, industry LLM.
We find that 42.5\% of a random sample of 3,692 of the Published Poems contain lines with exact matches in Dolma, using the What's In My Big Data (WIMBD) toolkit \citep{elazar2023s}.
When attempting to replicate the LLM probes, we found that both OpenAI and Anthropic models now refuse such completion queries.

\section{Linearization Methods and Evaluation}
\label{section-evaluation}

Crucially, for our dataset of Published Poems, it is not sufficient to scrape a poem's webpage; that webpage (its HTML or screenshot) must be parsed and converted into text that isolates the poem of interest while preserving whitespace formatting.
To transform the scraped data to poem texts, we test a series of linearization and image-to-text systems.

\subsection{Methods}

\paragraph{HTML to Text}

We compare a series of linearization systems for converting the scraped HTML to text.
These include resiliparse \cite{bevendorff:2018}, trafilatura \cite{barbaresi-2021-trafilatura}, and jusText.\footnote{https://github.com/miso-belica/jusText}
These tools have been used in production of pretraining datasets such as the Pile \cite{gao2020pile800gbdatasetdiverse}, Dolma \cite{dolma}, the RefinedWeb Dataset \cite{penedo2023refinedwebdatasetfalconllm}, OpenWebMath  \cite{paster2023openwebmathopendatasethighquality}, and DataComp-LM \cite{li2025datacomplmsearchgenerationtraining}.
Where possible, we have prioritized using default settings to simulate the processes leading to real pretraining datasets and the real effects of these parsers on poetry data.\footnote{\textbf{Resiliparse}: \texttt{preserve\_formatting = True}, \texttt{main\_content = True}, \texttt{list\_bullets = True}, \texttt{alt\_texts = False}, \texttt{links = False}, \texttt{form\_fields = False}, \texttt{noscript = False, comments = True}, \texttt{skip\_elements = None} (replicated from the code used to create the Dolma dataset \cite{dolma}); \textbf{Trafilatura}: \texttt{include\_comments = False}, \texttt{include\_links = False}, \texttt{include\_tables = False}, \texttt{no\_fallback = False}, \texttt{favor\_precision = False}, \texttt{favor\_recall = False}, \texttt{include\_formatting = False} (NB: changing \texttt{include\_formatting} to \texttt{True} does not alter results for poetry data) (replicated from the code used for DataTrove \citep{penedo2024datatrove}); \textbf{jusText}: \texttt{justext.get\_stoplist('English')},\texttt{length\_low = 0}, \texttt{length\_high = 100000}, \texttt{stopwords\_low = 0.0}, \texttt{stopwords\_high = 1.0},  \texttt{max\_link\_density = 1.0}, \texttt{no\_headings = False} (NB: stopwords are given but not used because of the thresholds) (attempted reasonable defaults).}
We run each pipeline over parts of the scraped webpages that isolate the \texttt{<div>} elements that contain the poems.
Importantly, these three methods operate on the scraped HTML without accounting for CSS styling or Javascript.
As noted by Clueweb \cite{overwijk2022clueweb2210billionweb}, ``the HTML alone provides a partial view of a web page,'' and so this is a limitation of these methods.

\paragraph{WISP-ify}
As a baseline comparison, we develop a custom HTML-to-text pipeline, WISP-ify, that accounts for the Poetry Foundation’s diverse formatting practices. 
The site uses whitespace in a variety of ways to convey lineation, stanza breaks, and visual emphasis. 
Our parser accommodates four major styles, including line- and stanza-level \texttt{<div>} elements, single paragraphs with \texttt{<br>} line breaks, multiple \texttt{<p>} tags for stanzas, and center-aligned lines. 
We convert left-margin spacing from inline CSS styles (e.g., \texttt{margin-left}) into corresponding plain-text indentation. We also normalize typographic features such as ligatures, small caps, and rare Unicode space characters. While our approach captures many of the site’s formatting conventions, others remain unsupported, and the site’s underlying structure may evolve in ways that challenge long-term reproducibility.

\paragraph{Image to Text}
HTML-only linearizers are constrained by an inability to capture CSS/Javascript styling essential to preserving whitespace. 
We capture ``screenshots'' of the poem using Playwright\footnote{\texttt{https://playwright.dev}} browser automation over Poetry Foundation HTML content, specifically targeting $.poem-body$ elements with fixed 1920x1080 viewport rendering. 
Each poem is thus converted to a PNG file.
We pass the image to three instruction-following multimodal models (o3, claude-sonnet-4, gemini-2.5-pro) prompting them to return whitespace-preserving text blocks (Appendix \ref{appendix-ocr-prompt}).

\input{tables/human_eval}

\subsection{Human Evaluation Setup} 
We introduce WISP-Bench to evaluate whitespace preservation fidelity across various linearization methods. 
WISP-Bench consists of a three-tiered set of pass-or-fail unit-tests, each of which asks: \textit{Given the ground truth image of the poem, does the linearized text accurately capture a specific whitespace property?} 
This design was inspired by olmOCR \cite{poznanski2025olmocr}, and the unit test guidelines are shown in Appendix \ref{appendix-wisp-bench}.

We curate a dataset of 76 poems that include whitespace features.\footnote{line break: 76 poems, vertical: 70, prefix: 64, internal: 40} 
For each of our seven linearization methods, the four authors evaluate the linearized text against the corresponding poem ``screenshot'' on WISP-Bench unit tests, such that each poem-method instance has at least two annotations.
As this is very difficult task, requiring careful attention to small changes in whitespace, we resolve disagreements by always preferring labels marking mistakes.

 We report pass rates across different WISP types for each method. For aggregation, we use four  scores to capture different aspects of the method: (1) Macro: Mean of pass-rates across WISP types, treating each type equally; (2) Weighted: Weighted mean of type pass-rates, biased towards the most frequent whitespace types; (3) Composite: A custom heuristic that penalizes OCR errors (see Appendix \ref{appendix-wisp-bench}), and (4) Pure: Pass rate across all annotations that have no OCR errors at all.

\subsection{How well do different linearization methods capture whitespace patterns?}
\label{section:results-linearization-methods}

Results of our human evaluation are shown in Table \ref{table:wisp-performance}. The relatively low macro scores highlight the complexity of preserving whitespace via linearization methods across modality, a facet not explicitly captured in traditional LLM-OCR benchmarks \cite{fu2025ocrbenchv2improvedbenchmark}. We note that specialized tools parsing HTML structure outperform general extraction methods, particularly due to the presence of hallucinated whitespace in LLMs (high OCR error-rate). 
We also note that LLMs exhibit similar strengths (line breaks) and weaknesses (prefix/internal spacing), possibly reflecting the common nature of their pretraining practices.

Figure~\ref{figure:barplot-linearization-comparison} in Appendix~\ref{appendix-linearization-comparison} shows prefix and internal whitespace patterns for three methods: resiliparse, trafilatura, and our custom pipeline (see \S\ref{subsection:data-published-poems}).
We find no meaningful difference between our pipeline and resiliparse, but trafilatura removes all \colorbox{prefix-color}{prefix} spacing.
We find that resiliparse very closely approximates our custom pipeline, while trafilatura and jusText mostly fail to preserve non-standard whitespace usages.
Trafilatura in particular is an interesting case, as it is designed to preserve whitespace only in detected code blocks.\footnote{\url{https://github.com/adbar/trafilatura/blob/master/trafilatura/htmlprocessing.py\#L324}}

We show an extended example in Figure~\ref{figure:parse-comparison} in the Appendix, which highlights the challenges in choosing a linearization pipeline. 
None of the tested HTML to text methods fully reproduce the spatial arrangement that can be seen on the Poetry Foundation website, though some methods come closer than others.
Ultimately, the spatial arrangement is a visual problem, which our findings underscore, and this will need to be handled using multimodal models in future work.

In our following analyses, we rely on texts generated with resiliparse, as it is a popular tool and had reasonable performance on WISP\feather{}-Bench (especially for \colorbox{prefix-color}{prefix} and \colorbox{internal-color}{internal} whitespace).

\begin{figure}[t]               
    \centering\includegraphics[width=0.8\linewidth]{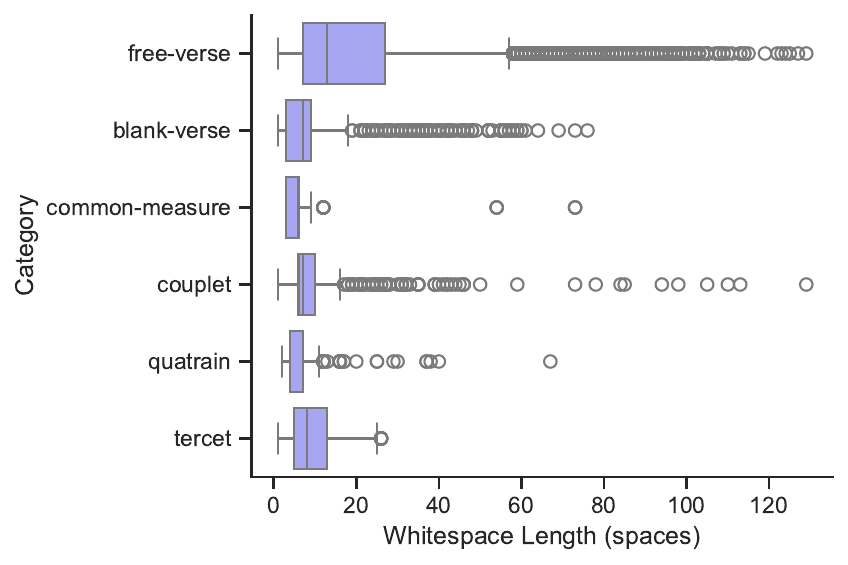}
    \caption{\colorbox{prefix-color}{Prefix} whitespace lengths, Published Poems.}
    \label{figure:boxplot-prefix}
\end{figure}

\begin{figure}[t]
    \centering\includegraphics[width=0.9\linewidth]{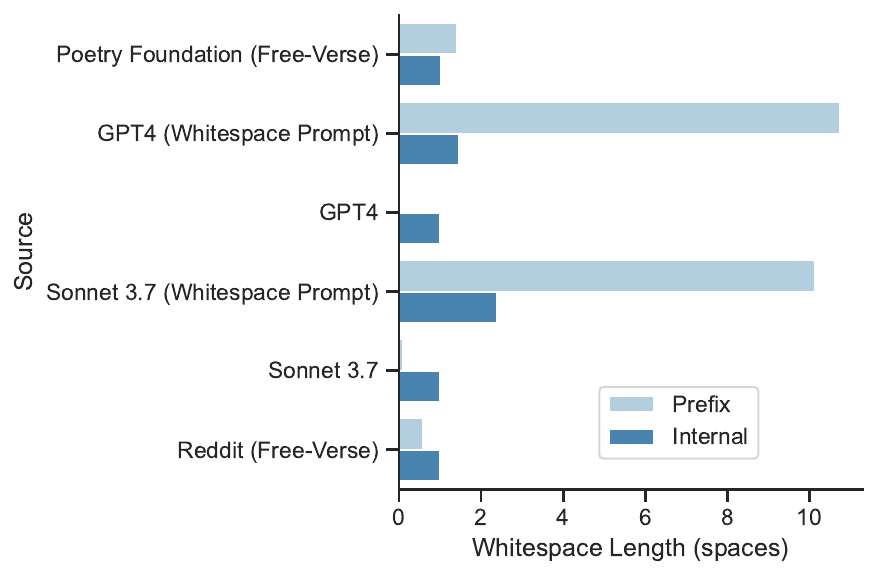}
    \caption{Comparison of \colorbox{prefix-color}{prefix} and \colorbox{internal-color}{internal} mean whitespace usage across the source datasets. To ensure a fair comparison, we compare the generated poems (which are almost all free-verse) only to free-verse poems from Poetry Foundation (as tagged on the website) and Reddit (as predicted using a prompt; see \S\ref{subsection:data-reddit}).}
    \label{figure:barplot-source-comparison-internal-prefix}
\end{figure}

\begin{figure}[t]
    \centering\includegraphics[width=\linewidth]{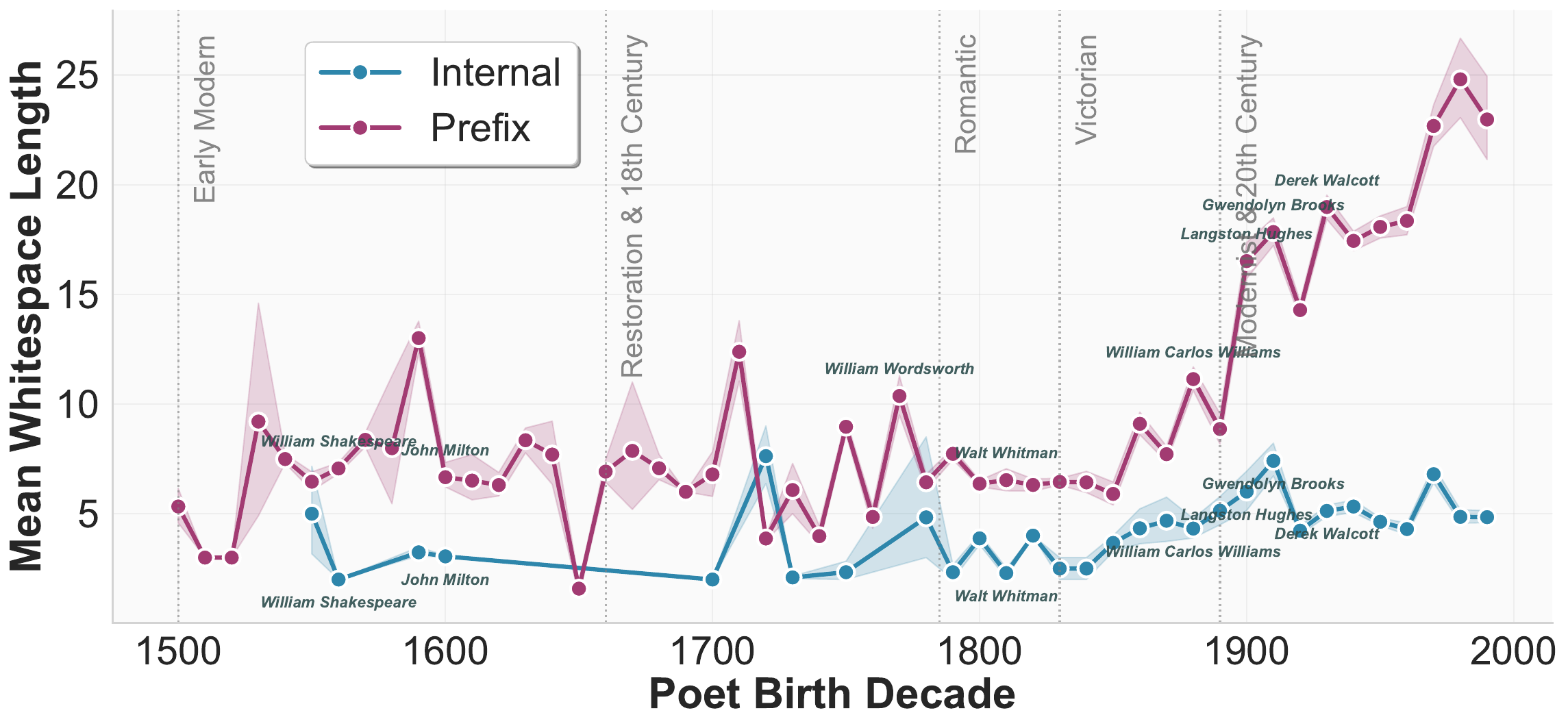}
    \caption{\colorbox{prefix-color}{Prefix} and \colorbox{internal-color}{internal} whitespace usages over time. The y-axis shows the mean number of spaces included in the whitespace, for all non-standard whitespace usages (we excluded non-standard usages from the denominator to highlight increasingly bold usages over time). Shaded areas show 95\% confidence intervals, and period lines are based on the \textit{Norton Anthology of English Literature}, 11th edition.}
    \label{figure:lineplot-prefix-internal-over-time}
\end{figure}

\begin{figure}[t]
    \centering\includegraphics[width=0.85\linewidth]{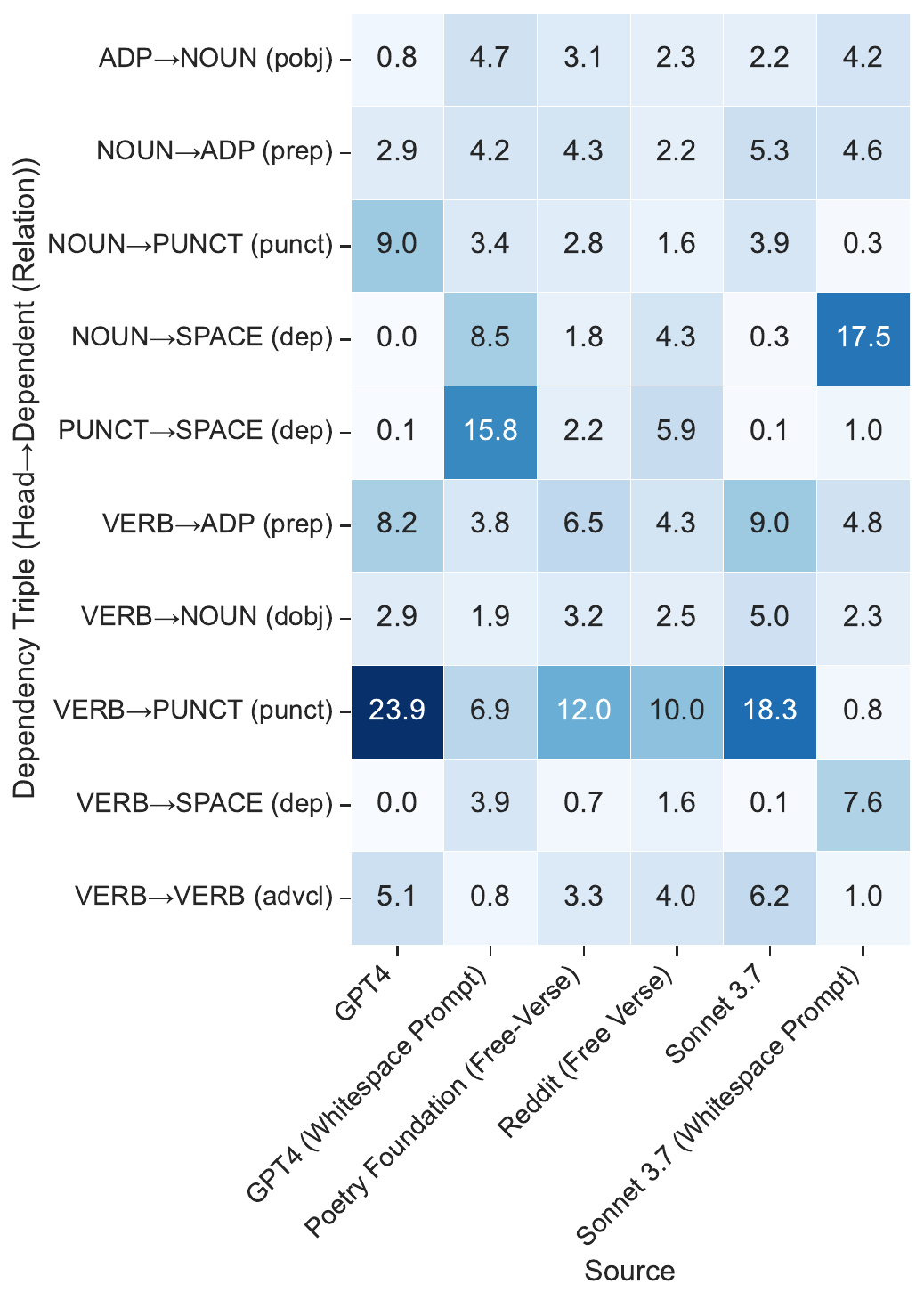}
    \caption{Comparison of most frequent dependency triples that span \colorbox{break-color}{line breaks} across the source datasets.}
    \label{figure:heatmap-spanning-triple-comparisons-by-source}
\end{figure}

\input{tables/ranked_categories_prefix}

\input{tables/ranked_punctuation}

\section{Analysis}
\label{section:analysis}

Due to space and feasibility constraints, we focus our computational analysis in this paper on three categories:\colorbox{break-color}{line breaks}, \colorbox{prefix-color}{prefix spacing}, and \colorbox{internal-color}{internal spacing}.
Our experiments explore whitespace as a stylistic choice and compare whitespace across data sources, tags, and forms.

\subsection{How does whitespace vary over published, unpublished, and generated poems?}
\label{across-source}

We find that published poems include more creative or non-standard whitespace (especially \colorbox{prefix-color}{prefix spacing}) than poems on Reddit, at least when written in free verse (Figure~\ref{figure:barplot-source-comparison-internal-prefix}), possibly due to formatting difficulties on Reddit.
When prompted to generate a poem with no explicit mention of whitespace in the prompt, GPT-4 and Sonnet 3.7 almost never produce poems with non-standard prefix spacing.
However, they are clearly capable of producing whitespace-heavy poems.
When we use our whitespace specific prompt, the models generate poems with more prefix whitespace on average than the Poetry Foundation poems.

In Figure~\ref{figure:heatmap-spanning-triple-comparisons-by-source}, we observe different kinds of dependency triples occurring at \colorbox{break-color}{line breaks} across datasets. 
The most common triple across published poems, unpublished human poems, and the default LLM prompt is \texttt{VERB -> PUNCT}.
This suggests that enjambment often occurs after complete syntactic units, especially after verbs followed by punctuation.
It reflects a poetic style that uses enjambment for rhythm, pacing, or breath, not necessarily to break grammar mid-thought.
It may also reflect how parsers attach punctuation to verbs, making this a common dependency pair in any sentence-final line---especially in free verse.

By contrast, we find that LLMs with the explicit whitespace prompt most often produce \texttt{NOUN -> SPACE} or \texttt{PUNCT -> SPACE} triples that span across line breaks.
In other words, generated poems not only use internal and prefix spacing more frequently, they also use whitespace differently (with different types of line break enjambements) than human-written published or unpublished poems.

\subsection{How does whitespace vary by poetic form?}

Across all forms, free verse contains the widest variation of whitespace and the most \colorbox{prefix-color}{prefix space} on average (Figure~\ref{figure:boxplot-prefix}), while couplets include the most \colorbox{internal-color}{internal space} on average (Figure \ref{figure:boxplot-internal}).

As in \S\ref{across-source}, \texttt{VERB -> PUNCT} is the most common dependency triple spanning a line break for all forms in published poems (Figure~\ref{figure:barplot-triple-proportions-resiliparse}).

Table~\ref{table:ranked-punctuation} shows differences in the punctuation preceding \colorbox{break-color}{line breaks} across the different forms.
Commas are the most common punctuation at line end across all the forms.
However, colons (``:'') and semicolons (``;'') are more likely to appear at line end than elsewhere in the line, especially for couplets and common measure.
Significantly, free verse poems overall have less frequent punctuation at line breaks, reflecting the creative spatial organization that is representative of this form.

\subsection{Has whitespace usage changed over time?}

Figure~\ref{figure:lineplot-prefix-internal-over-time} suggests that poets have steadily used more  whitespace over the last 500 years. 
We represent poems temporally by the decade of the author's birth year.
Birth year has been used in prior work to examine innovation in literary and cultural change \citep{griebel2024locating}.
We do not control for the number of data points per poet, as poets can and do adapt their stylistic choices over time, and such changes are themselves of literary interest.
For any instance of \colorbox{prefix-color}{prefix spacing} or non-standard \colorbox{internal-color}{internal space}, we find the mean number of spaces.
We do so to highlight bold and idiosyncratic choices.
We see that the size of such whitespace usage is increasing, especially in the 20th century, and especially for prefix spacing.

\subsection{How does whitespace vary by topic?}

To characterize the kinds of poems with the highest and lowest whitespace usage, we first determine which poems include whitespace lengths above the 75th percentile (calculated using all whitespace lengths from every poem and every tag).
We then find the proportion of poems assigned to each tag (manual labels applied by Poetry Foundation) that are in this high whitespace usage category. 
Tables \ref{table-tags-prefix} and \ref{table-tags-internal} show the top tags for \colorbox{prefix-color}{prefix} and \colorbox{internal-color}{internal} whitespace, with example poets whose poem(s) have the highest/lowest whitespace usage among all poems with that tag.
We only show tags assigned to at least $N=100$ poems.
As expected, we see tags for traditional forms like ``Sonnet'' ranked lowest for whitespace usage, while we see tags for modern topics like ``Gender-Sexuality'' and physicalities like ``The Body'' ranked highest.

\section{Discussion}

Paying closer attention to whitespace opens up new avenues for computational literary and cultural analysis, enabling macro-level studies of how poetic form and visual layout have changed over time. 
In the twentieth century, advancements in printing and typesetting technologies gave poets greater freedom to experiment spatially, and whitespace has become integral to meaning-making, rhythm, and reader engagement. 
Our findings confirm this scholarly narrative and demonstrate how
researchers can explore innovation across historical periods, literary movements, or national traditions.

But we find that distinguishing deliberate whitespace from formatting artifact noise is extremely challenging when a poem has been transferred through various mediums (manuscript to print, print to print, print to digital) and formats (HTML/image/text), due to the inherent typographic inconsistencies of diverse rendering engines, font metrics, character encoding, and responsive layouts.
We have also observed, in the dataset of Reddit poems, the importance of different platforms, whose affordances can shape poets' choices. 
Given the rarity of standardized ground truth (and the difficulties of adjudicating a ``ground truth'' in this setting, where even archival scholarship might not produce an obvious ranking of one version over another), the development of accurate whitespace linearization methods is crucial for preserving authorial intent---even if mediated by different formats.

More ambitiously, modeling whitespace at this scale might lead to advancements in computational tools for poetry scholarship and digital literary preservation.
Multimodal LLM tools could assist in or even partially automate the labor-intensive process of encoding poetic texts using systems like the Text Encoding Initiative (TEI).

However, we caution that such systems must always keep domain experts in the loop, as encoding poetry in TEI is a fundamentally interpretive act that involves annotating specific elements of texts for particular goals \citep{Flanders}. 
While some affordances of TEI would be difficult to productively automate, accurately capturing whitespace could cut down significantly on the labor involved in reproducing the layouts of poetic texts \citep{micir}. 

For LLM data collectors and model builders, poetry provides an instructive test case.
While much attention has been given to the formatting of programming and mathematical inputs \citep{paster2023openwebmathopendatasethighquality}, whitespace in poetry is more idiosyncratic, and we do not know of existing off-the-shelf linearization systems that are designed to handle poetry.
As prior work has argued \citep{walsh-etal-2024-sonnet}, poetry is a popular generation task and a ``lightning rod'' for public imagination around artificial intelligence capabilities, and is worthy of research attention.
Practically, we recommend resiliparse as a baseline linearization method for scraped poetry data.
However, none of our tested methods faithfully captured all whitespace usage as shown visually on the Poetry Foundation website.
Future work will need to tackle the CSS and other styling outside of the HTML and incorporate more advanced multimodal and vision model pipelines.

\section{Conclusion}
Our work introduces a whitespace typology for poetry, which we use to investigate how 4k poets from the Poetry Foundation have linguistically and syntactically used whitespace in 19.4k poems across 500 years. We compare this usage to 51.4k LLM-generated poems and 11.9k unpublished poems posted in the subreddit \texttt{r/OCPoetry} and discuss differences in their distribution. We also discuss the impact of different linearization methods on our results. Finally, we release 2.8k public-domain poems with preserved whitespace formatting to facilitate future work.

\section{Limitations}

Our whitespace and linguistic analysis is limited to English-language poems in the Roman script and may not translate to poetry in other languages or scripts. Similarly, our representation of poets across time is also restricted to their digital presence on the Poetry Foundation, and hence our conclusions are not truly representative of all English poets of any given time.
These poems over-represent poets from the North American region.
In addition, LLMs can ``memorize'' training data, which often contains copyright-protected literary work. 
During generation, these models may bear resemblance to the original poems despite our explicit prompt instruction to not reuse original text. 

Of course, poems are present in pretraining datasets not only through scraped web data but also through book data \cite{chang2023speakmemoryarchaeologybooks}.
We observe this even in our scraped poems, which when searched for in Dolma, as described in \S\ref{subsection:data-published-poems}, return the most hits from a single domain from Google Books.
It is likely that poem texts taken from books also suffer from whitespace issues due to OCR and other errors, but we leave this investigation to future work.

\section{Ethical Considerations}

The literary community of poets, readers, editors, and publishers faces significant challenges due to recent advances in LLMs and synthetically generated poetry that mimics human verse with unprecedented fidelity on the syntactic level \cite{porter2024ai}. A poem is a human artistic endeavor that captures the agency, expression, reflection, and communal meaning-making of the poet's lived experiences. Synthetically generated poems lack this sense of meaning; literary magazines and publishers aiming to filter out such synthetically generated submissions are struggling with the complexity of the task and the increased load of submissions.\footnote{https://clarkesworldmagazine.com/clarke\_04\_23/} As \textit{Rattle Magazine} succinctly puts it, ``Poetry is a tool for expanding the human spirit, which means poems should be written by humans.''\footnote{https://rattle.com/page/submissions/} We encourage future work in the computational study of poetry to use WISP\feather{} for building effective analysis and detection tools to help the literary community, but acknowledge that our work can also be misused for generative optimizations which hinder such causes instead.

We used Claude (Anthropic) to assist in the generation of boilerplate code used to process the data and produce early versions of figures. All code was tested and most code was re-written after using Claude for brainstorming.

\section{Acknowledgments}

This work was supported by Doing AI Differently, a joint initiative of The Alan Turing Institute and University of Edinburgh, funded by Arts and Humanities Research Council (AHRC-UKRI).
We would like to thank to Kyle Lo and Luca Soldaini (for advice and feedback) and Lynn Cherny, Amit Chaudhary, Barry Haddow, and Mithun Hunsur (for sharing key references).
We also thank the Simpson Center for the Humanities at the University of Washington for their general support of digital humanities scholarship. 
Thank you to the Poetry Foundation, and thank you to the poets Shankar Narayan, Bill Carty, and Jeanine Walker for their inspiration.

\bibliography{MAIN}

\clearpage
\newpage

\appendix
\section{Appendix}
\label{sec:appendix}

We show examples of poems with complex whitespace usages and provide further results in this Appendix.

\begin{figure}[h]
    \centering
    \includegraphics[width=9cm]{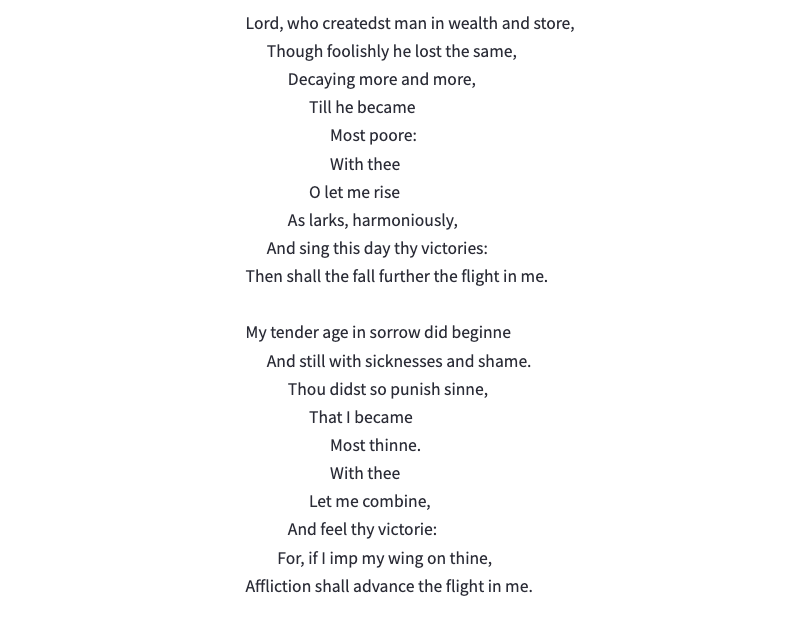}
    \caption{``\href{https://www.poetryfoundation.org/poems/44361/easter-wings}{[Easter Wings]}'' by George Herbert (1593—1633), from the Poetry Foundation.}
    \label{fig:easter-wings}
\end{figure}

\begin{figure}[h]
    \centering
    \includegraphics[width=9cm]{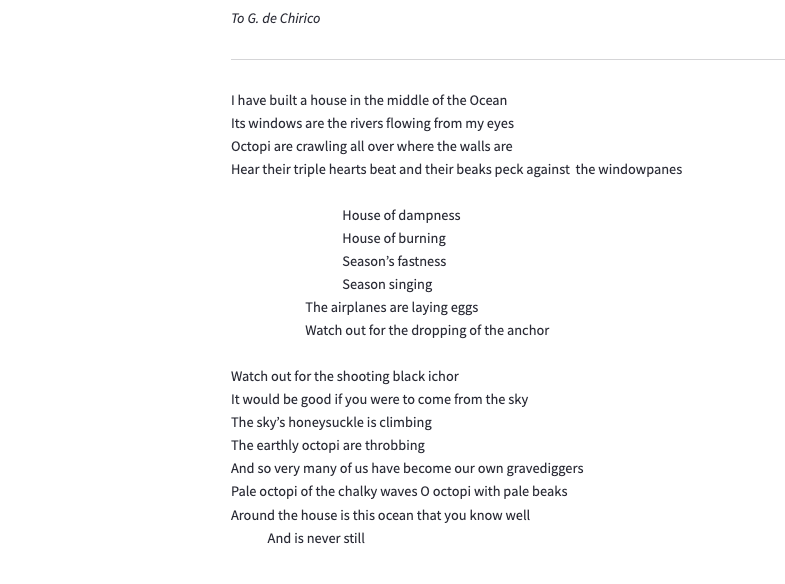}
    \caption{``\href{https://www.poetryfoundation.org/poetrymagazine/poems/58343/ocean-of-earth}{[Ocean of Earth]}'' by Guillaume Apollinaire (1880-1918), translated from French by Ron Padgett}
    \label{fig:ocean-of-earth}
\end{figure}

\begin{figure}[h]
    \centering
    \includegraphics[width=9cm]{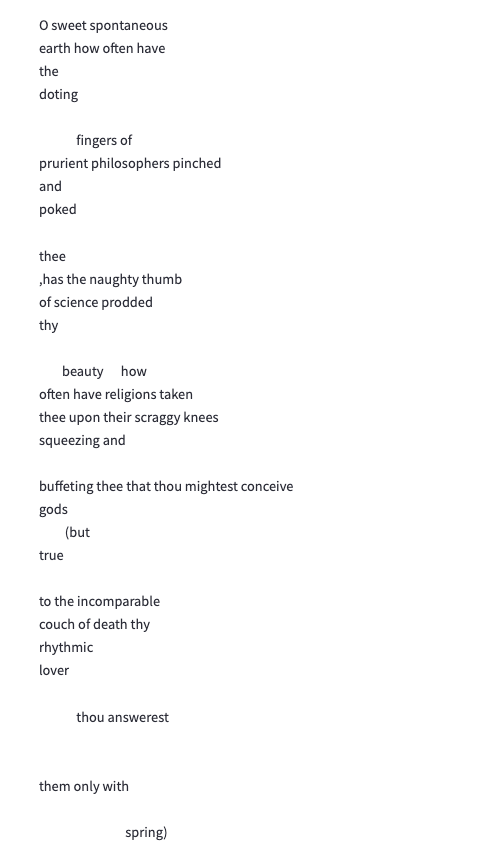}
    \caption{``\href{https://www.poetryfoundation.org/poems/148505/o-sweet-spontaneous-5bf31932ce110}{[O sweet spontaneous]}'' (\copyright 1923) by E.E. Cummings, from the Poetry Foundation.}
    \label{fig:o-sweet}
\end{figure}

\subsection{Comparison of HTML to Text Methods}

\begin{figure*}[h]

    \begin{subfigure}[t]{.3\linewidth}
        \centering\includegraphics[width=4cm]{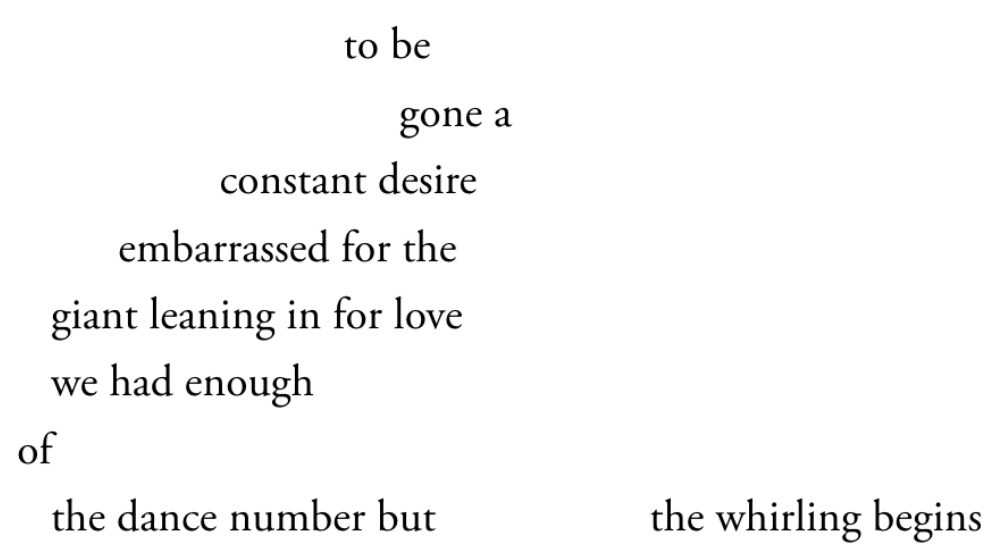}
        \caption{Poetry Foundation}
    \end{subfigure}
    \begin{subfigure}[t]{.3\linewidth}
        \centering\includegraphics[width=4cm]{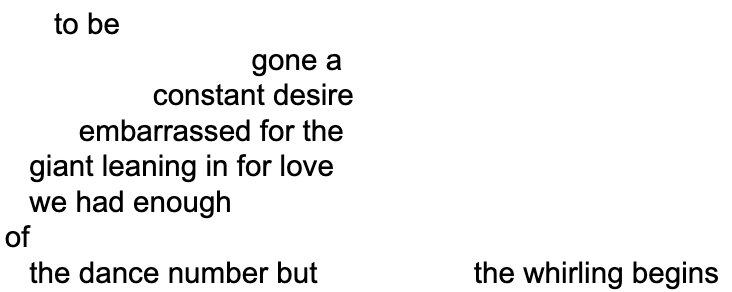}
        \caption{Resiliparse}
    \end{subfigure}
    \begin{subfigure}[t]{.3\linewidth}
        \centering\includegraphics[width=4cm]{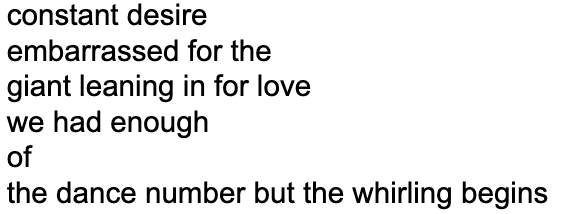}
        \caption{Trafilatura}
    \end{subfigure}
    \begin{subfigure}[t]{.3\linewidth}
        \centering\includegraphics[width=4cm]{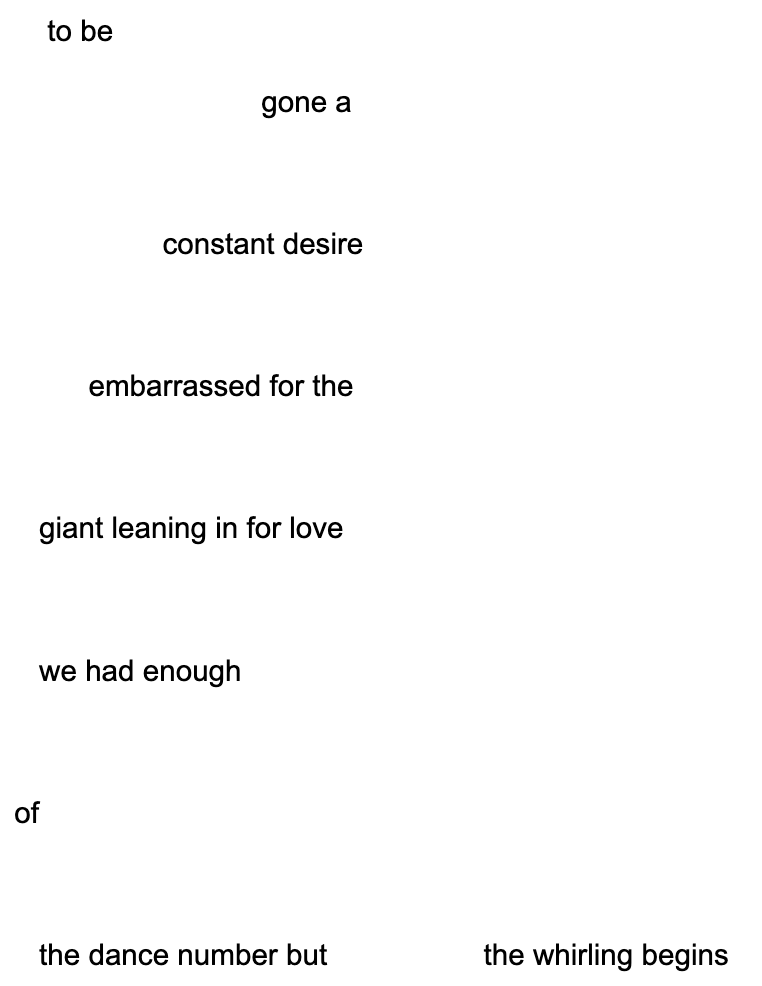}
        \caption{BeautifulSoup}
    \end{subfigure}
    \begin{subfigure}[t]{.3\linewidth}
        \centering\includegraphics[width=4cm]{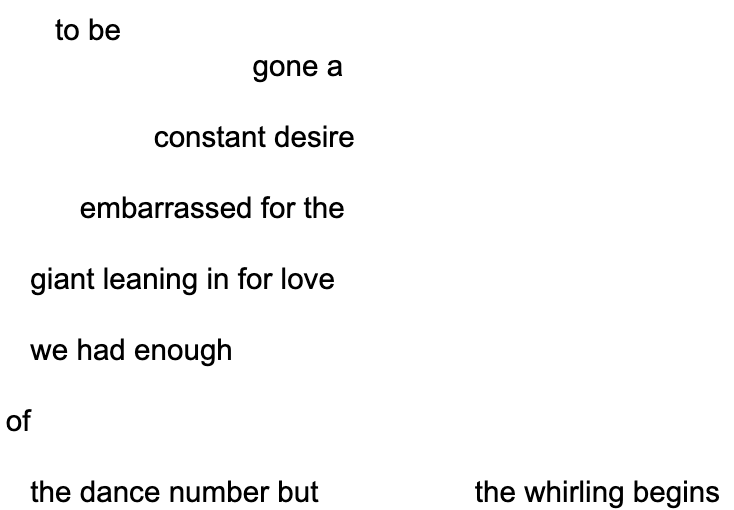}
        \caption{HTML2text}
    \end{subfigure}
    \begin{subfigure}[t]{.3\linewidth}
        \centering\includegraphics[width=4cm]{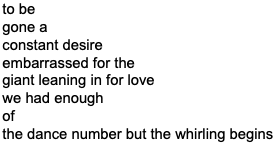}
        \caption{jusText}
    \end{subfigure}

    \caption{Comparisons of the opening lines of the poem ``Mars.1'' (2016) by CAConrad across different HTML to text methods.}
    \label{figure:parse-comparison}
\end{figure*}

\subsection{Whitespace, Part-of-Speech, and Dependency Triples by Poetic Form}

\begin{figure}[H]
    \centering\includegraphics[width=0.8\linewidth]{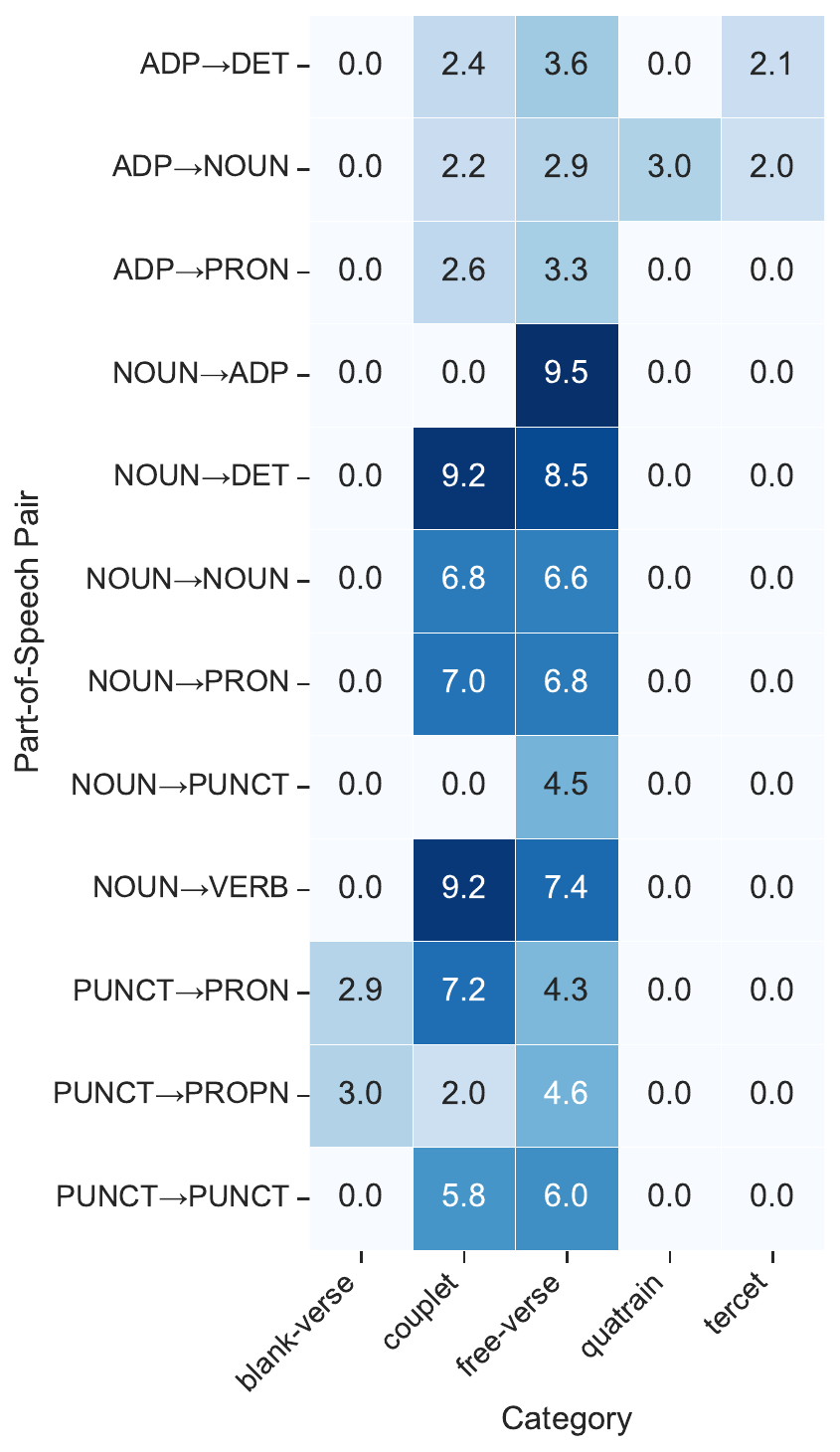}
    \caption{The average \colorbox{internal-color}{internal} whitespace length between pairs of POS tags for the Published Poems parsed using resiliparse.} 
    \label{figure:heatmap-spanning-triple-lengths-resiliparse}
\end{figure}

\begin{figure}[H]
     \centering\includegraphics[width=0.9\linewidth]{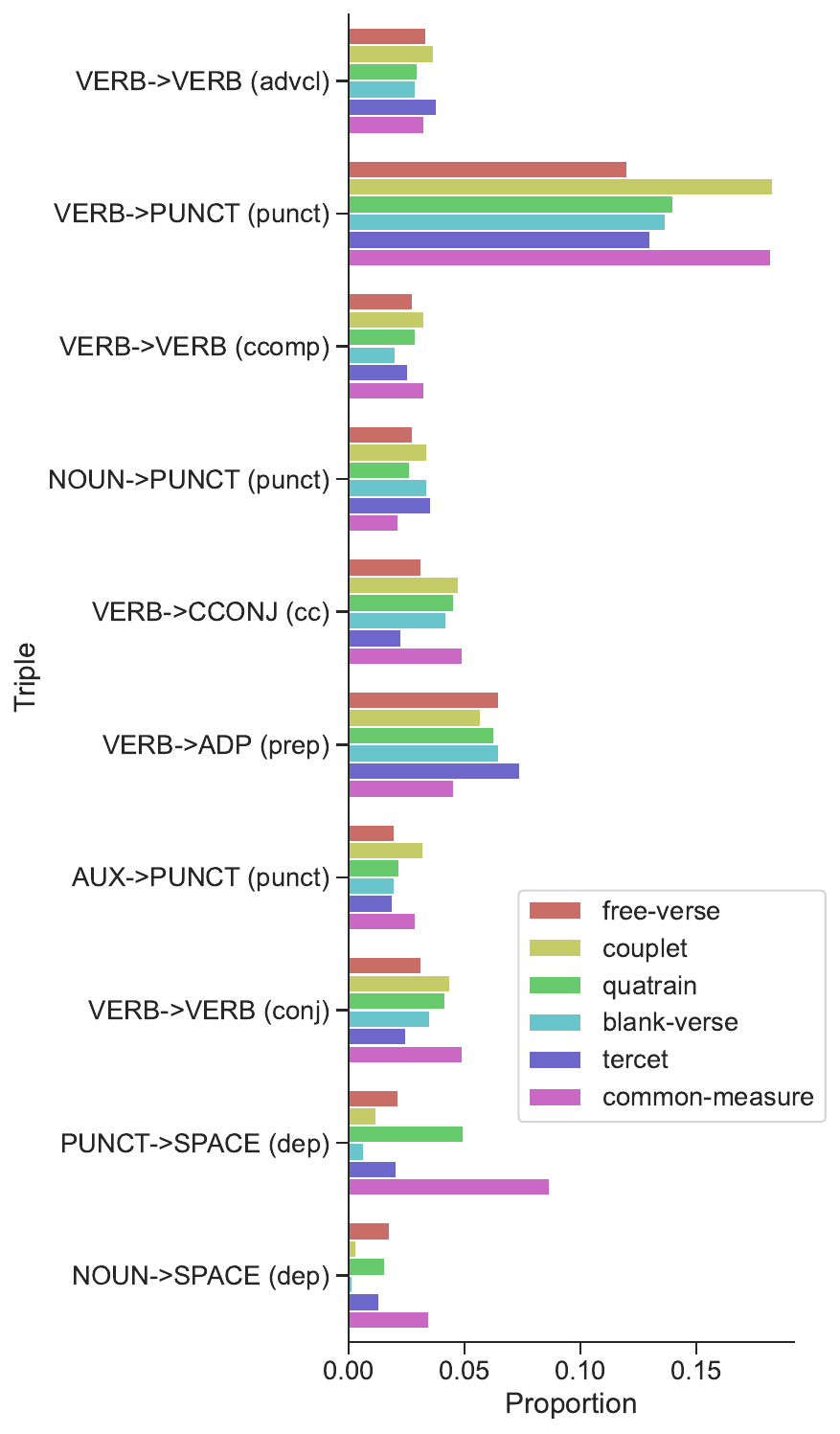}
    \caption{The proportions of the most common dependency triples (\texttt{head POS->dependent POS (relation type)}) that span across \colorbox{break-color}{line breaks} for the Published Poems parsed using resiliparse. These proportions represent only lines \textit{not} ending at a sentence boundary.}
    \label{figure:barplot-triple-proportions-resiliparse}
\end{figure}

\subsection{Linearization Comparison}
\label{appendix-linearization-comparison}

\begin{figure}[H]
\centering\includegraphics[width=0.9\linewidth]{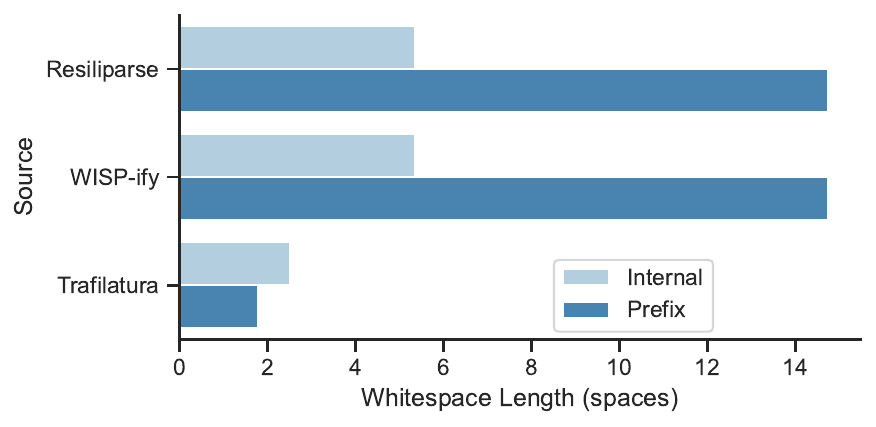}
    \caption{Comparison of \colorbox{prefix-color}{prefix} and \colorbox{internal-color}{internal} mean whitespace lengths across three HTML to text methods, including our custom pipeline described in \S\ref{subsection:data-published-poems}. 
    These results are normalized only by the total number of non-standard usages, not the total number of lines or internal spaces, to highlight differences.}
    \label{figure:barplot-linearization-comparison}
\end{figure}

\subsection{Forms and Whitespace}

\begin{figure}[H]
    \centering\includegraphics[width=0.7\linewidth]{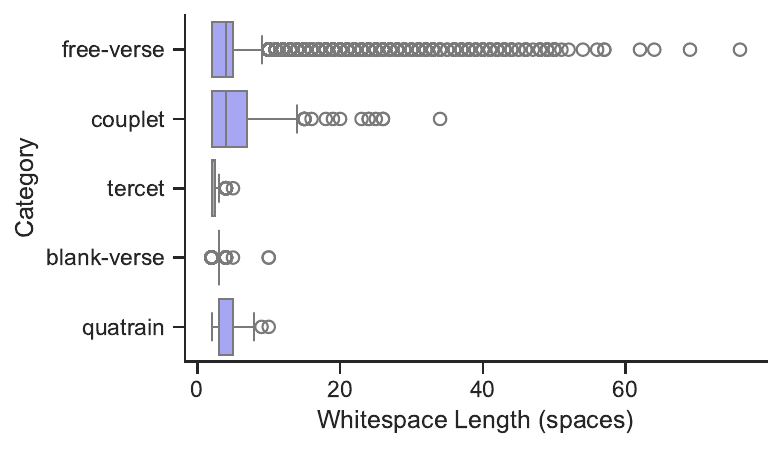}

    \caption{Lengths of \colorbox{internal-color}{internal} whitespace usages for Published Poems.}
    \label{figure:boxplot-internal}
\end{figure}

\subsection{Tags and Whitespace}

\input{tables/ranked_categories_internal}

\section{Poem Generation Prompt}
\label{appendix-poem-generation-prompt}

\begin{tcolorbox}[colbacktitle=table-title-color,
title=Poem Generation Prompt (Whitespace),coltitle=black,
fonttitle=\bfseries,colback=table-background-color]
\hspace{-2mm}
    \footnotesize
    
    I'm very interested interested in how you use whitespace for poetry data. Could you display your capabilities by writing three new poems inspired by the themes of the poem ``\texttt{poem\_title}'' by \texttt{poet\_name}.
    
    \vspace{0.2cm}
    
    I want your new poems to use whitespace creatively, in ways that are appropriate for each poem. Each poem should use whitespace differently. This could include enjambment, vertical spacing between lines, prefix spacing before the first word in a line, or line-internal spacing between or within words.

    \vspace{0.2cm}
    
    Do not use any text from the original poem. Print your new poems inside <poem></poem> tags and then provide explanations of your whitespace usage inside <explanation></explanation> tags. Make sure your output is in plain text and do not include a title.
\end{tcolorbox}

\section{WISP\feather{}-Bench}
\label{appendix-wisp-bench}

\subsection{A Three Tiered Benchmark}

Given the ``spectrum of correctness'' of whitespace fidelity, WISP-Bench has three hierarchical tiers of evaluation:

\begin{itemize}[leftmargin=*]
\item \textbf{Presence Match} Structural Fidelity - do the basic spatial elements (line break/prefix/internal/vertical spacing) exist where they should?

\item \textbf{Fuzzy Match} Relational Fidelity - are the proportional relationships between whitespace elements preserved? For example, if two consecutive whitespace elements in the image are 2 and 4 spaces, and their respective textual counterparts are 4 and 8 spaces, relative spatial presence is said to be preserved.

\item \textbf{Exact Match} Absolute Fidelity - has the precise visual layout and appearance been preserved? While this is difficult to evaluate due to the challenge of transforming pixels to characters, this requires exact correspondence of structure.
\end{itemize}

\subsection{Unit Tests in the Benchmark}

\begin{enumerate}[leftmargin=*]

\item \textbf{Line Break Test (Presence)}\\
\begin{tabularx}{\linewidth}{@{}p{0.22\linewidth}@{\hspace{0.015\linewidth}}Y@{}}
\textit{Question:} & Does the text capture line breaks where they should be? \\
\textit{Check If:} & The first and last words of the printed line N (between two \textbackslash ns) in the text match their corresponding positions in the image, for all N.
\end{tabularx}

\item \textbf{Prefix Space Tests}
\begin{itemize}[leftmargin=*]
    \item \textbf{2a. Prefix (Presence)}\\
    \begin{tabularx}{\linewidth}{@{}p{0.22\linewidth}@{\hspace{0.015\linewidth}}Y@{}}
    \textit{Question:} & Is indentation preserved at all? \\
    \textit{Check If:} & There is at least one instance of a prefix whitespace being preserved.
    \end{tabularx}
    
    \item \textbf{2b. Prefix (Fuzzy)}\\
    \begin{tabularx}{\linewidth}{@{}p{0.22\linewidth}@{\hspace{0.015\linewidth}}Y@{}}
    \textit{Question:} & Are relative indentation levels preserved? \\
    \textit{Check If:} & Ranking of indentation depths matches (line A more indented than B), if there's more than 1 prefix whitespace line in the poem.
    \end{tabularx}
    
    \item \textbf{2c. Prefix (Exact)}\\
    \begin{tabularx}{\linewidth}{@{}p{0.22\linewidth}@{\hspace{0.015\linewidth}}Y@{}}
    \textit{Question:} & Are exact indentation levels preserved? \\
    \textit{Check If:} & Number of leading spaces/tabs matches within tolerance ($\pm$\,1 space). Does this pass the eye test---does the prefix spacing \emph{look} perfectly preserved?
    \end{tabularx}
\end{itemize}

\item \textbf{Internal Space Tests}
\begin{itemize}[leftmargin=*]
    \item \textbf{3a. Internal (Presence)}\\
    \begin{tabularx}{\linewidth}{@{}p{0.22\linewidth}@{\hspace{0.015\linewidth}}Y@{}}
    \textit{Question:} & Is extra spacing between words preserved? \\
    \textit{Check If:} & There is at least one instance of an internal whitespace being preserved.
    \end{tabularx}
    
    \item \textbf{3b. Internal (Fuzzy)}\\
    \begin{tabularx}{\linewidth}{@{}p{0.22\linewidth}@{\hspace{0.015\linewidth}}Y@{}}
    \textit{Question:} & Are relative internal spacing levels preserved? \\
    \textit{Check If:} & Ranking of internal space depths is preserved (word pair AB more indented than CD), if there's $>$\,1 internal whitespace word pair in the poem.
    \end{tabularx}
    
    \item \textbf{3c. Internal (Exact)}\\
    \begin{tabularx}{\linewidth}{@{}p{0.22\linewidth}@{\hspace{0.015\linewidth}}Y@{}}
    \textit{Question:} & Are exact internal spacing amounts preserved? \\
    \textit{Check If:} & The number of internal spaces matches within tolerance. Eye test---does the internal spacing \emph{look} right?
    \end{tabularx}
\end{itemize}

\item \textbf{Vertical Space Tests}
\begin{itemize}[leftmargin=*]
    \item \textbf{4a. Vertical Space (Presence)}\\
    \begin{tabularx}{\linewidth}{@{}p{0.22\linewidth}@{\hspace{0.015\linewidth}}Y@{}}
    \textit{Question:} & Is vertical spacing ($>$\,1 newline) preserved? \\
    \textit{Check If:} & There is at least one instance of 2 newline characters / 1 blank line present between lines.
    \end{tabularx}
    
    \item \textbf{4b. Vertical Space (Relative)}\\
    \begin{tabularx}{\linewidth}{@{}p{0.22\linewidth}@{\hspace{0.015\linewidth}}Y@{}}
    \textit{Question:} & Are relative vertical spacing levels preserved? \\
    \textit{Check If:} & Ranking of vertical space matches (line pair AB more separated than CD), if there's $>$\,1 vertical-space line pair in the poem.
    \end{tabularx}
    
    \item \textbf{4c. Vertical Space (Exact)}\\
    \begin{tabularx}{\linewidth}{@{}p{0.22\linewidth}@{\hspace{0.015\linewidth}}Y@{}}
    \textit{Question:} & Are exact vertical spacing amounts preserved? \\
    \textit{Check If:} & The number of newlines between the lines is preserved (no tolerance since newlines are conspicuous). Eye test: Do the new lines \emph{look} right?
    \end{tabularx}
\end{itemize}

\end{enumerate}

\vspace{0.5em}
\noindent\textbf{NOTE:} We have left out line\_lengths from the annotation due to challenges in devising unit tests for this type of whitespace usage. 

\subsection{Scoring Metrics}

Let $U$ denote the set of unit tests, $A_u$ the annotations containing unit test $u$, and $T_u$ true accepts for option $u$. Let annotation sets be partitioned as \textit{catastrophic}: $C$ (only OCR Error is labeled true, other tests are marked false); \textit{mixed}: $M$ (OCR Error is true, but there is at least one unit test that has passed); and \textit{pure}: $P$ (OCR Error is false).

\begin{description}
    \item[\textbf{Reliability Factor}] 
    \begin{equation}
    R = 1 - \left(\frac{|C|}{|A|} + 0.5 \times \frac{|M|}{|A|}\right)
    \end{equation}

    \item[\textbf{Macro Score}] 
    \begin{equation}
    \text{Macro} = \frac{1}{|U|} \sum_{u \in U} \frac{|T_u|}{|A_u|} \times 100
    \end{equation}
    
    \item[\textbf{Weighted Macro Score}] 
    \begin{equation}
    \text{Weighted} = \frac{\sum_{u \in U} |T_u|}{\sum_{u \in U} |A_u|} \times 100
    \end{equation}
    
    \item[\textbf{Composite Score}] 
    \begin{equation}
    \text{Composite} = \text{Macro} \times R
    \end{equation}
    
    \item[\textbf{Pure Score}] 
    \begin{equation}
    \text{Pure} = \frac{1}{|U|} \sum_{u \in U} \frac{|T_u \cap P|}{|A_u \cap P|} \times 100
    \end{equation}
\end{description}

\section{OCR Transcription Prompt for Multimodal LLMs}
\label{appendix-ocr-prompt}

\begin{lstlisting}[basicstyle=\ttfamily\scriptsize,breaklines=true]
SYSTEM_PROMPT = """
## Objective:
Convert the poem image into plain text with exact preservation of its visual layout (spacing, alignment, and line breaks). Prioritize fidelity to the image structure and visual layout over standard formatting. Your task is purely transcription with layout preservation. Do not interpret, explain, or modify the text.

## Formatting Guidelines:
 Here are some guidelines to help with edge cases:
 - Use □ for unreadable characters
 - Ignore all typographical formatting like *italics*, **bold**, 'underline', or strikethrough. Transcribe only the text and its spacing.
 - **DO NOT** auto-wrap long lines. If a line in the image is very long, it must be preserved as a single line in the output, as line breaks (enjambment) are a poetic device.
 - In case of columnar poems, maintain the column structure using spaces in each row to preserve visual structure. Make sure the rows are aligned correctly across all columns.
 - If text is centered or right-aligned, replicate the alignment using spaces so it visually matches the image.
 - If there are gaps within a line (e.g., scattered words or concrete poetry effects), preserve the spacing exactly as in the image.
 - Alignment/indentation: Align word positions precisely with reference lines above/below, preserving exact indentation levels between successive lines. For instance, if the word 'foo' in the second line is spaced in a way that the 'f' aligned with the 'b' in the word 'bar' in the previous line in the image, then it should be reflected similarly in the text. 
 - In case of newlines/vertical spacing, preserve the exact number of newlines and vertical gaps as seen in the image.
 - In case of concrete poems / scattered poems, the visual layout of the image is a part of the semantics of the poem. Capture it faithfully as possible with spaces.
 - Accurately represent all non-English and special characters (é, ç, ß, etc.) using their exact Unicode code points. Do not use approximations (e.g., don't replace é with e).
- Use appropriate single Unicode characters for superscripts/subscripts (e.g., ², ₁). 
 - For erasure/blackout poetry, transcribe only the visible text and use spaces to represent the blacked-out areas, preserving the position of the remaining words.
 - In case of page numbers and sections breaks, preserve the layout and spacing exactly as it appears in the image.
 - For superscript/subscript/interpolation of multiple characters, use the appropriate Unicode characters (e.g., ² for superscript 2, ₁ for subscript 1) and ensure they are placed correctly in relation to the surrounding text.
 - In case of rotated/upside-down characters, use the corresponding Unicode character wherever possible.
 - **Ligatures:** Decompose typographic ligatures into their constituent characters (e.g., transcribe 'ﬁ' as 'fi', 'ﬂ' as 'fl', and 'æ' as 'ae').

## Prioritization in Cases of Conflict
All guidelines serve the primary objective, but if rules appear to conflict, follow this strict priority order:
 - **Most Important** Global Layout > Local Spacing: Prioritize the overall "shape" and structure. If maintaining the exact space count between two words causes a column or a centered block to become misaligned, always prioritize the global alignment (the column's starting position, the text's center point) over the exact local space count.
 - **Specific Poem Types > General Rules:** Rules for specific types (like `erasure poetry`) **always override** general formatting rules (like `ignore all... strikethrough`).
 - Visual Alignment > Semantic Characters: The highest priority is to make the text output *look* like the image. Instructions to use specific Unicode characters (like `²` or `₁`) or to decompose ligatures (like `ﬁ` to `fi`) must **be ignored** if following them would alter the character count or width in a way that breaks the poem's visual alignment. In such a conflict, transcribe the characters *exactly as needed to hold the visual shape*, even if it means using standard characters (like `f` and `i` separately) to match the layout.

## Output Format:
- Output must consist of exactly one fenced code block containing only the transcription. Do not include explanations, labels, or commentary outside the block.
- Output must be valid UTF-8 text using only ASCII spaces (U+0020) and standard line breaks (LF: U+000A) for whitespace.
"""
\end{lstlisting}

\end{document}

%% file: COMMANDS.tex
\definecolor{maria-color}{HTML}{7881F2}
\definecolor{melanie-color}{HTML}{ac78f2}
\definecolor{harsh-color}{HTML}{006400}

\usepackage{booktabs}
\usepackage{array}
\newcolumntype{P}[1]{>{\centering\arraybackslash}p{#1}}
\newcolumntype{L}[1]{>{\raggedright\let\newline\\\arraybackslash\hspace{0pt}}p{#1}}
\newcolumntype{R}[1]{>{\raggedleft\arraybackslash}p{#1}}

\usepackage{graphicx,subcaption}

\usepackage{pgf}
\usepackage{colortbl}

\definecolor{color1}{HTML}{93003a}
\definecolor{color2}{HTML}{cf3759}
\definecolor{color3}{HTML}{f4777f}
\definecolor{color4}{HTML}{ffbcaf}
\definecolor{color5}{HTML}{ffffe0}
\definecolor{color6}{HTML}{a5d5d8}
\definecolor{color7}{HTML}{73a2c6}
\definecolor{color8}{HTML}{4771b2}
\definecolor{color9}{HTML}{00429d}

\usepackage{array}

\newcommand*{\opacity}{30}
\newcommand*{\minval}{-1}
\newcommand*{\maxval}{1}

\newcommand{\gradient}[1]{
    \ifdimcomp{#1pt}{>}{\maxval pt}{#1}{
        \ifdimcomp{#1pt}{<}{\minval pt}{#1}{
            \pgfmathparse{int(round(8*(#1/(\maxval-\minval))-(\minval*(8/(\maxval-\minval)))))}
            \xdef\tempa{\pgfmathresult}
            \ifcase\tempa
                \cellcolor{color1!\opacity} #1\or
                \cellcolor{color2!\opacity} #1\or
                \cellcolor{color3!\opacity} #1\or
                \cellcolor{color4!\opacity} #1\or
                \cellcolor{color5!\opacity} #1\or
                \cellcolor{color6!\opacity} #1\or
                \cellcolor{color7!\opacity} #1\or
                \cellcolor{color8!\opacity} #1\or
                \cellcolor{color9!\opacity} #1
            \fi
    }}
}

\newcommand{\gradientspecificity}[1]{
    \ifnum#1>0
        \cellcolor{green!10} #1
    \else
        \cellcolor{red!10} #1
    \fi
}

\usepackage{xcolor}
\definecolor{break-color}{HTML}{d5fad4}
\definecolor{prefix-color}{HTML}{f9d4fa}
\definecolor{internal-color}{HTML}{cff5f5}
\definecolor{vertical-color}{HTML}{e2cff5}
\definecolor{length-color}{HTML}{f5e2cf}

\usepackage{graphicx}
\newcommand{\feather}[1]{\includegraphics[height=1em]{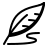}}
\newcommand{\reddit}[1]{\includegraphics[height=1em]{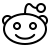}}
\newcommand{\book}[1]{\includegraphics[height=1em]{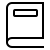}}
\newcommand{\sparkle}[1]{\includegraphics[height=1em]{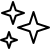}}

\usepackage{tcolorbox}
\usepackage{float}
\usepackage{enumitem}
\definecolor{table-title-color}{HTML}{93c8ff}
\definecolor{table-background-color}{HTML}{e8f3ff}

%% file: tables/typology.tex
\begin{table}[t]
    \centering
    \small
    \begin{tabular}{lp{2cm}p{2.5cm}}
    
        \toprule
        \textbf{Category} & \textbf{Sub-Category} & \textbf{Definition} \\
        
        \midrule
        
        \colorbox{break-color}{\texttt{line break}} & standard (!e) & breaks at sentence boundary \\
        \colorbox{break-color}{\texttt{line break}} & lexical (e) & a word is split across two lines \\
        \colorbox{break-color}{\texttt{line break}} & clausal (e) & a clause (noun and verb) is split across two lines \\
        \colorbox{break-color}{\texttt{line break}} & phrasal (e) & a phrase (e.g., adjective and noun) is split across lines \\
        
        \midrule
         
        \colorbox{prefix-color}{\texttt{prefix}} & standard & no indent  \\
        \colorbox{prefix-color}{\texttt{prefix}} & standard indent & repeated indent that aligns with a form \\
        \colorbox{prefix-color}{\texttt{prefix}} & non-standard & all other indents \\
        
        \midrule

        \colorbox{internal-color}{\texttt{internal}} & standard & single white space between tokens  \\
        \colorbox{internal-color}{\texttt{internal}} & non-standard & multiple spaces between words or within a word \\
        \midrule
         
        \colorbox{vertical-color}{\texttt{vertical}} & standard & a single newline character \\
        \colorbox{vertical-color}{\texttt{vertical}} & standard stanza & two newline characters between stanzas \\
        \colorbox{vertical-color}{\texttt{vertical}} & non-standard & multiple newline characters not at stanza boundaries \\

        \midrule
         
        \colorbox{length-color}{\texttt{line length}} & standard & uniform line lengths across the poem \\
        \colorbox{length-color}{\texttt{line length}} & non-standard & non-uniform line lengths \\
        
        \bottomrule
    \end{tabular}
    \caption{Our typology of whitespace usage in poems. (e: enjambed, !e: not enjambed)
    }
    \label{table:typology}
\end{table}

%% file: tables/data.tex
\begin{table*}[t]
    \centering
    \small
    \begin{tabular}{p{3cm}p{3cm}p{2cm}p{2cm}p{2cm}}
        \toprule
        \textbf{Category} & \textbf{Source} & \textbf{Poem Count} & \textbf{Mean \newline Line Count} & \textbf{Most Common Form} \\
        \midrule
        \textbf{\book{} Published Poems} & Poetry Foundation &  19,457 & 38.1 & sonnet \\[0.1cm]
        \textbf{\reddit{} Unpublished Poems} & \texttt{r/OCPoetry} (Reddit) & 11,984 & 26.5 &  free-verse \\[0.1cm]
        \textbf{\sparkle{} Generated Poems} & GPT-4 (OpenAI) &   12,838 & 11.9 &   free-verse  \\
        & GPT-4 (OpenAI) & 12,645 & 10.5 & free-verse  \\
        & Sonnet 3.7 (Anthropic) & 12,988 & 12.5 & free-verse  \\
        & Sonnet 3.7 (Anthropic) & 12,987 & 11.3 & free-verse  \\
        \bottomrule
    \end{tabular}
    \caption{The three datasets that we collect. Vocabulary density represents unique token counts divided by the total token counts. Small differences in poem counts across generated poems are due to generation/parsing errors.}
    \label{table:data}
\end{table*}

%% file: tables/human_eval.tex
\begin{table*}[t]
    \centering
    \scriptsize
    \begin{tabular}{l|cccc|ccccc}
        \toprule
        \textbf{Method} & \textbf{Macro} & \textbf{Weighted} & \textbf{Composite} & \textbf{Pure} & \textbf{PREFIX} & \textbf{INTERNAL} & \textbf{LINE\_BREAKS} & \textbf{VERTICAL} &
        \textbf{OCR-ERROR}  \\
        \midrule
        Resiliparse & \textbf{51.66} & \textbf{52.22} & \textbf{49.28} & 53.79 & \textbf{48.44} & \textbf{45.83} & 63.16 & \textbf{71.90} & 7.89 \\
        WISP-ify & 50.44 & 51.04 & 43.80 & \textbf{55.88} & 45.31 & \textbf{45.00} & 63.16 & \textbf{70.95} & 17.11 \\
        jusText & 3.35 & 4.15 & 2.86 & 3.41 & 0.00 & 0.00 & 34.21 & 0.00 & 15.79 \\
        trafilatura & 3.11 & 3.86 & 2.95 & 3.28 & 0.00 & 0.00 & 34.21 & 0.00 & \textbf{5.26} \\
        \midrule
        \textit{Claude Sonnet 4} & 45.48 & 46.13 & 35.41 & \textbf{56.35} & 38.00 & 42.16 & 72.13 & 56.55 & 31.15 \\
        \textit{Gemini 2.5 Pro} & 45.08 & 45.74 & 41.47 & 46.38 & 33.85 & 42.74 & \textbf{78.67} & 57.14 & 16.0 \\
        \textit{o3} & 42.80 & 43.77 & 33.79 & 48.56 & 33.33 & 37.50 & 65.79 & 57.14 & 31.58 \\
        \bottomrule
    \end{tabular}
    \caption{Human evaluation of linearization method performance across WISP whitespace types. Italicized methods are image-to-text, the rest are HTML-to-text. Scores representing best performance $\pm0.1$ are bolded.}
    \label{table:wisp-performance}
\end{table*}

%% file: tables/ranked_categories_prefix.tex
\begin{table}[h]
\scriptsize
\centering
\begin{tabular}{lccl}
\toprule
\multicolumn{4}{c}{Highest \textbf{Prefix} Whitespace Usage} \\
\midrule
Tag & N & Proportion & Example Poet \\
\midrule
Gay-Lesbian-Queer & 184 & 0.418 & Wendy Videlock \\
Persona & 145 & 0.388 & Gottfried Benn \\
Epigraph & 144 & 0.370 & Nick Carbó \\
Gender-Sexuality & 788 & 0.359 & Wendy Videlock \\
Stars-Planets-Heavens & 320 & 0.347 & Amy E. Sklansky \\
Popular Culture & 467 & 0.345 & Allen Ginsberg \\
Free Verse & 4881 & 0.345 & Elizabeth Bishop \\
\midrule
\multicolumn{4}{c}{Lowest \textbf{Prefix} Whitespace Usage} \\
\midrule
Tag & N & Proportion & Example Poet \\
\midrule
Common Measure & 122 & 0.007 & Elinor Wylie \\
Ballad & 117 & 0.018 & [...] Montagu \\
Funerals & 108 & 0.030 & Jean Nordhaus \\
Quatrain & 151 & 0.031 & Adam Zagajewski \\
Verse Forms & 912 & 0.037 & Deborah Paredez \\
Sonnet & 622 & 0.046 & Deborah Paredez \\
Animals-1 & 115 & 0.048 & \textit{anonymous} \\
\bottomrule
\end{tabular}
\caption{Tags with highest/lowest \colorbox{prefix-color}{prefix} whitespace.}
\label{table-tags-prefix}
\end{table}

%% file: tables/ranked_punctuation.tex
\begin{table*}[t]
    \centering
    \footnotesize
    \rowcolors{2}{violet!10}{white}
    \begin{tabular}{lp{6cm}p{6cm}}
        \toprule
        \textbf{Form} & \textbf{Most Common Punctuation at Line End \newline (Per Total Lines)} & \textbf{Most Likely Punctuation at Line End \newline (Per Punctuation Token Usage)} \\
        \midrule
\textbf{free-verse} 
    & \textbf{\colorbox{gray!20}{\texttt{,}} (12.6\%)} \colorbox{gray!20}{\texttt{.}} (10.1\%) \colorbox{gray!20}{\texttt{—}} (1.1\%) \colorbox{gray!20}{\texttt{?}} (0.9\%) 
    & \textbf{\colorbox{gray!20}{\texttt{;}} (41.1\%)} \colorbox{gray!20}{\texttt{?}} (33.1\%) \colorbox{gray!20}{\texttt{:}} (33.1\%) \colorbox{gray!20}{\texttt{.}} (32.8\%) \\[0.1cm]

\textbf{couplet} 
    & \textbf{\colorbox{gray!20}{\texttt{,}} (26.0\%)} \colorbox{gray!20}{\texttt{.}} (10.9\%) \colorbox{gray!20}{\texttt{;}} (7.8\%) \colorbox{gray!20}{\texttt{:}} (3.6\%) 
    & \textbf{\colorbox{gray!20}{\texttt{;}} (79.1\%)} \colorbox{gray!20}{\texttt{:}} (72.5\%) \colorbox{gray!20}{\texttt{?}} (52.6\%) \colorbox{gray!20}{\texttt{)}} (44.7\%) \\[0.1cm]

\textbf{quatrain} 
    & \textbf{\colorbox{gray!20}{\texttt{,}} (18.5\%)} \colorbox{gray!20}{\texttt{.}} (9.0\%) \colorbox{gray!20}{\texttt{;}} (2.5\%) \colorbox{gray!20}{\texttt{—}} (1.4\%) 
    & \textbf{\colorbox{gray!20}{\texttt{;}} (58.7\%)} \colorbox{gray!20}{\texttt{?}} (38.0\%) \colorbox{gray!20}{\texttt{:}} (36.6\%) \colorbox{gray!20}{\texttt{,}} (36.4\%) \\[0.1cm]

\textbf{blank-verse} 
    & \textbf{\colorbox{gray!20}{\texttt{,}} (25.6\%)} \colorbox{gray!20}{\texttt{.}} (8.4\%) \colorbox{gray!20}{\texttt{;}} (3.7\%) \colorbox{gray!20}{\texttt{:}} (2.1\%) 
    & \textbf{\colorbox{gray!20}{\texttt{)}} (48.0\%)} \colorbox{gray!20}{\texttt{.}} (43.2\%) \colorbox{gray!20}{\texttt{—}} (42.8\%) \colorbox{gray!20}{\texttt{?}} (41.9\%) \\[0.1cm]

\textbf{tercet} 
    & \textbf{\colorbox{gray!20}{\texttt{,}} (10.9\%)} \colorbox{gray!20}{\texttt{.}} (9.2\%) \colorbox{gray!20}{\texttt{:}} (0.7\%) \colorbox{gray!20}{\texttt{?}} (0.6\%) 
    & \textbf{\colorbox{gray!20}{\texttt{:}} (25.0\%)} \colorbox{gray!20}{\texttt{.}} (24.9\%) \colorbox{gray!20}{\texttt{,}} (21.3\%) \colorbox{gray!20}{\texttt{?}} (20.2\%) \\[0.1cm]

\textbf{common-measure} 
    & \textbf{\colorbox{gray!20}{\texttt{,}} (29.2\%)} \colorbox{gray!20}{\texttt{;}} (10.9\%) \colorbox{gray!20}{\texttt{.}} (6.6\%) \colorbox{gray!20}{\texttt{!}} (1.5\%) 
    & \textbf{\colorbox{gray!20}{\texttt{;}} (89.4\%)} \colorbox{gray!20}{\texttt{,}} (51.6\%) \colorbox{gray!20}{\texttt{!}} (30.5\%) \colorbox{gray!20}{\texttt{.}} (29.2\%) \\[0.1cm]

        \bottomrule
    \end{tabular}
    \caption{The most common punctuation at \colorbox{break-color}{line breaks} across poetic forms. Left: proportion of lines ending in a punctuation token, normalized by the total number of lines. Right: proportion of a punctuation token ($N>=100$) appearing at the end of a line, normalized by that token’s total usage in any place in a poem.}
    \label{table:ranked-punctuation}
\end{table*}

%% file: tables/ranked_categories_internal.tex
\begin{table}[H]
\scriptsize
\centering
\begin{tabular}{lccl}
\toprule
\multicolumn{4}{c}{Highest \textbf{Internal} Whitespace Usage} \\
\midrule
Tag & N & Proportion & Example Poet \\
\midrule
Ghosts-the-Supernatural & 163 & 0.453 & Ching-In Chen \\
Gender-Sexuality & 788 & 0.373 & May Swenson \\
Refrain & 162 & 0.347 & Adam O. Davis \\
Series-Sequence & 271 & 0.326 & Toi Derricotte \\
Grief & 1840 & 0.323 & Terisa Siagatonu \\
Theater-Dance & 130 & 0.322 & Penelope Shuttle \\
The Body & 1737 & 0.311 & Toi Derricotte \\
\midrule
\multicolumn{4}{c}{Lowest \textbf{Internal} Whitespace Usage} \\
\midrule
Tag & N & Proportion & Example Poet \\
\midrule
Common Measure & 122 & 0.000 & Robert W. Service \\
Valentine's Day & 119 & 0.000 & Sir Philip Sidney \\
Blank Verse & 235 & 0.006 & Robert Pinsky \\
Tercet & 121 & 0.006 & Tom Sleigh \\
Funerals & 108 & 0.008 & Jean Nordhaus \\
Simile & 113 & 0.009 & [...] Anne Finch \\
Rhymed Stanza & 1702 & 0.027 & Edmund Spenser \\
\bottomrule
\end{tabular}
\caption{Tags with highest/lowest \colorbox{internal-color}{internal} whitespace.}
\label{table-tags-internal}
\end{table}